\pdfoutput=1

\documentclass[11pt]{article}

\usepackage{acl}
\usepackage{hyperref}
\usepackage{times}
\usepackage{latexsym}
\usepackage{multirow}
\usepackage{booktabs}
\usepackage{microtype}
\usepackage{multirow}
\usepackage{latexsym}
\usepackage{booktabs}
\usepackage{courier}
\usepackage{amsmath}
\usepackage{enumitem}
\usepackage{booktabs} 
\usepackage{siunitx} 
\usepackage{subfig}
\usepackage{placeins}
\usepackage{tikz}
\usepackage{hyperref}
\DeclareMathOperator*{\argmax}{argmax}

\usepackage{xspace}

\newcommand{\ie}{i.e.,\xspace}
\newcommand{\eg}{e.g.,\xspace}

\usepackage[T1]{fontenc}

\usepackage[utf8]{inputenc}

\usepackage{microtype}

\usepackage{inconsolata}
\usepackage{tcolorbox}

\usepackage{pifont}
\title{Edinburgh Clinical NLP at SemEval-2024 Task 2:\\Fine-tune your model unless you have access to GPT-4}

\author{
Aryo Pradipta Gema\textsuperscript{1}\thanks{These authors contributed equally to this work.} \quad Giwon Hong\textsuperscript{1}\footnotemark[1] \quad Pasquale Minervini\textsuperscript{1} \quad Luke Daines\textsuperscript{2} \quad Beatrice Alex\textsuperscript{3,4} \\
\textsuperscript{1}School of Informatics, University of Edinburgh \quad 
\textsuperscript{2}Usher Institute, University of Edinburgh\\
\textsuperscript{3}Edinburgh Futures Institute, University of Edinburgh\\
\textsuperscript{4}School of Literatures, Languages and Cultures, University of Edinburgh\\
\texttt{ \{aryo.gema, giwon.hong, luke.daines, p.minervini, b.alex\}@ed.ac.uk}
}

\begin{document}
\maketitle
\begin{abstract}
    The NLI4CT task assesses Natural Language Inference systems in predicting whether hypotheses entail or contradict evidence from Clinical Trial Reports.
    In this study, we evaluate various Large Language Models (LLMs) with multiple strategies, including Chain-of-Thought, In-Context Learning, and Parameter-Efficient Fine-Tuning (PEFT).
    We propose a PEFT method to improve the consistency of LLMs by merging adapters that were fine-tuned separately using triplet and language modelling objectives.
    We found that merging the two PEFT adapters improves the F1 score (+0.0346) and consistency (+0.152) of the LLMs.
    However, our novel methods did not produce more accurate results than GPT-4 in terms of faithfulness and consistency.
    Averaging the three metrics, GPT-4 ranks joint-first in the competition with 0.8328.
    Finally, our contamination analysis with GPT-4 indicates that there was no test data leakage.\footnote{Our code is available at \url{https://github.com/EdinburghClinicalNLP/semeval_nli4ct}.}
\end{abstract}

\section{Introduction}
Extracting insights from Clinical Trial Reports (CTRs) is vital for advancing personalised medicine, yet manual analysis of these vast datasets is impractical. The Natural Language Inference for Clinical Trial Data (NLI4CT) task~\cite{jullien-etal-2024-semeval}\footnote{\url{https://sites.google.com/view/nli4ct/}} addresses this challenge by evaluating Natural Language Inference (NLI) systems' ability to understand and reason within this domain.

In this study, we evaluate various LLMs, such as LLaMA2~\cite{touvron2023bllama}, Mistral~\cite{jiang2023mistral}, MistralLite~\cite{MistralLite_2023}, and GPT-4~\cite{openai2023gpt4}. We employed prompting strategies like In-context Learning (ICL) and Chain-of-Thought (CoT) to improve their accuracy.
We also proposed a Parameter-Efficient Fine-Tuning (PEFT) method that merges independently fine-tuned adapters trained with distinct objectives, namely a triplet loss and a language modelling (LM) loss, to improve the consistency of the LLMs.

Our findings reveal that our novel PEFT method improves the F1 and consistency scores of the LLMs.
However, GPT-4 produces more accurate results than all of the models we considered, co-leading the competition leaderboard.
Although GPT-4 places fifth in the F1 score, its high faithfulness and consistency scores highlight its potential for a reliable prediction in the clinical domain.
Lastly, we conduct a contamination analysis of GPT-4 to check whether instances of the NLI4CT dataset were included in GPT-4's pre-training data.

\section{Background}


\begin{table*}[ht!]
\small
\centering
\begin{tabular}{lccccccccc}
\toprule
\multicolumn{1}{c}{\multirow{2}{*}{\textbf{Split}}} &
  \multicolumn{1}{c}{\multirow{2}{*}{\textbf{Total}}} &
  \multicolumn{2}{c}{\textbf{Sample Type}} &
  \multicolumn{4}{c}{\textbf{Section Type}} &
  \multicolumn{2}{c}{\textbf{Label Type}} \\
\cmidrule(lr){3-4}
\cmidrule(lr){5-8}
\cmidrule(lr){9-10}
\multicolumn{1}{c}{} &
  \multicolumn{1}{c}{} &
  \multicolumn{1}{c}{\textbf{Single}} &
  \multicolumn{1}{c}{\textbf{Comparison}} &
  \multicolumn{1}{c}{\textbf{Intervention}} &
  \multicolumn{1}{c}{\textbf{Eligibility}} &
  \multicolumn{1}{c}{\textbf{Results}} &
  \multicolumn{1}{c}{\textbf{Adverse Events}} &
  \multicolumn{1}{c}{\textbf{Ent.}} &
  \multicolumn{1}{c}{\textbf{Con.}} \\
\midrule
Train & 1,700 & 1,035 & 665   & 396   & 486   & 322   & 496   & 850   & 850   \\
Dev   & 200   & 140   & 60    & 36    & 56    & 56    & 52    & 100   & 100   \\
Test  & 5,500 & 2,553 & 2,947 & 1,542 & 1,419 & 1,235 & 1,304 & 1,841 & 3,659 \\
\bottomrule
\end{tabular}
\caption{Dataset statistics of each split, categorised by sample, section, and label types.}
\label{table:data_stat}
\end{table*}

\subsection{Task overview}
The NLI4CT task leverages a collection of CTRs and expert-annotated hypotheses. This iteration places a heightened emphasis on faithfulness (robustness to semantic changes) and consistency (stability against semantic preserving alterations). Aside from this focus, the composition of the dataset and the task objective remains identical to the previous iteration \cite{jullien-etal-2023-nli4ct, jullien-etal-2023-semeval}. Table \ref{table:data_stat} contains statistics for each data split, organised by sample, section, and label types.

\begin{description}[style=unboxed,leftmargin=0cm,noitemsep,topsep=0pt]
\item[Section Types] Each CTR consists of four  sections: ``Eligibility criteria'', ``Intervention'', ``Results'', and ``Adverse events''.
Hypotheses are sentences claiming information in a CTR section.

\item[Sample Types] The task presents two sample types: ``Single'' and ``Comparison''.
``Single'' samples provide all relevant evidence within one CTR, while ``Comparison'' samples require cross-referencing information from two CTRs.

\item[Task Objective] The task objective is to classify the relationship between hypotheses and corresponding CTR(s) as ``entailment'' or ``contradiction''.
``Entailment'' implies that the hypothesis is supported by the CTR(s), while a ``contradiction'' classification suggests inconsistency.

\end{description}


\subsection{Related work}

LLMs demonstrated promising results in the medical domain. For example, \citet{lievin2022can} conducted evaluations on LLMs, including Codex~\cite{chen2021evaluating} and InstructGPT~\cite{ouyang2022training} using zero-shot, few-shot, and CoT prompting.
These LLMs show comprehension of complex medical questions, recall of domain knowledge, and nontrivial reasoning.

Despite the increasing use of general LLMs, domain adaptive fine-tuning remains a prevailing approach in the medical domain~\cite{lehman2023we}.
As LLMs continue to grow in size, PEFT gains preference over full-parameter fine-tuning due to its resource efficiency.
\citet{gema2023parameter} proposed a two-stage PEFT framework, one for domain-adaptive pre-training and one for downstream fine-tuning, to adapt  LLaMA~\cite{touvron2023allama} to the clinical outcome prediction tasks.
Even though \citet{gema2023parameter} introduced the idea of combining multiple adapters, 
they did not explicitly merge the adapter weights.
\citet{chronopoulou2023adaptersoup} proposed AdapterSoup, which performs averaging of the weights of PEFT adapters trained on the same objective function and different domains to improve the model's performance.

Extending the adapter merging idea, we introduced a novel method to merge PEFT adapters that are trained on different training objectives: triplet loss and LM loss.
We compared this method with strategies without parameter fine-tuning, such as zero-shot inference, ICL, and CoT.

\section{System Overview}

We experimented with two strategies.
The first involved no fine-tuning, aiming to comprehend LLMs' inherent ability to solve clinical tasks.
The second employed our proposed PEFT method to improve the consistency of the model.
Both systems ingest CTR-hypothesis pairs, predicting the correct label one token at a time from left to right.
%

\subsection{Without Parameter Fine-tuning}
\label{sec:no_finetuning}
\begin{figure*}[ht]
    \centering
    \includegraphics[width=0.99\textwidth]{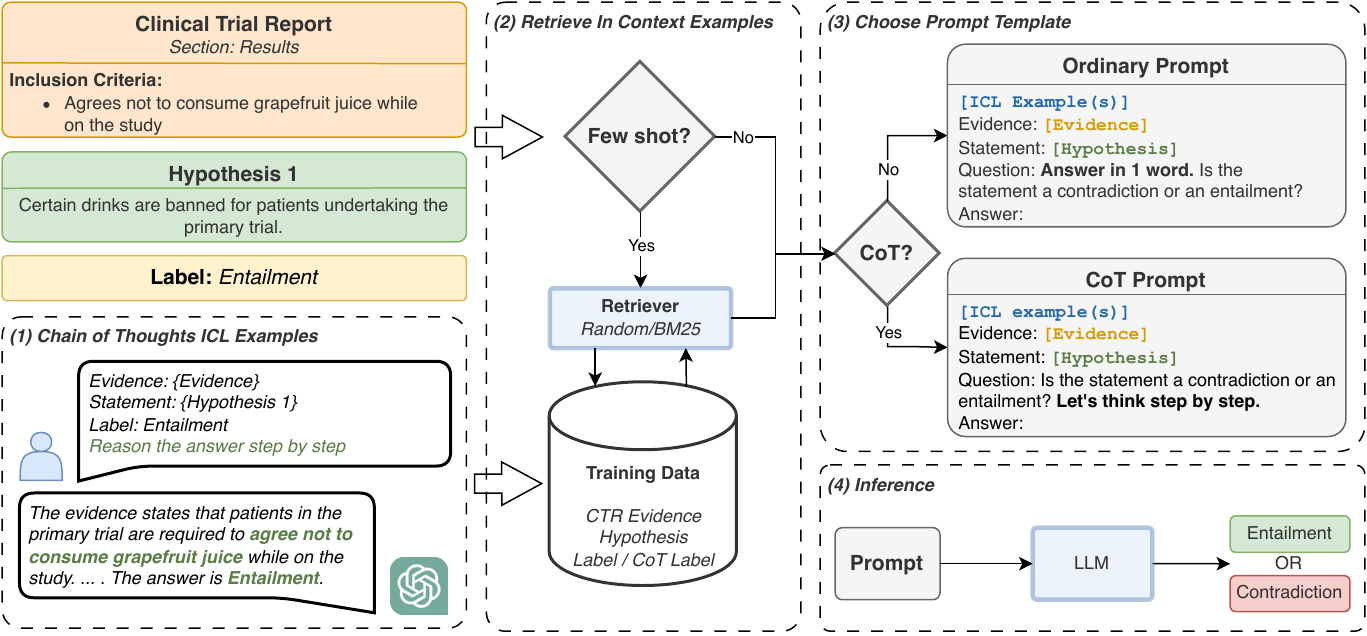}
    \caption{Our inference schema with multiple prompting strategies (without fine-tuning). For Chain-of-Thought examples, Natural Language Explanation was generated using ChatGPT~\cite{he2023using}.}
    \label{fig:no_fine_tuning_schema}
\end{figure*}

The system with no fine-tuning utilises the pretrained general LLMs for prediction.
We experimented with multiple prompting strategies:
\begin{description}[style=unboxed,leftmargin=0cm,noitemsep,topsep=0pt]
\item[Zero-shot] Employing the LLMs without any fine-tuning and examples.
\item[In-Context Learning (ICL)] Adapting the LLMs by providing examples of how to perform a task. Due to the maximum context length of the LLMs, we limit experiments to two examples (2-shot).
\item[Chain-of-Thought (CoT)] Prompting LLMs with a phrase (\eg~\emph{``Let's think step by step''})~\cite{kojima2022large}, encouraging a sequential reasoning.
\item[ICL + CoT] Adapting the LLMs with ICL examples that are augmented with reasoning steps.
\end{description}
%


\autoref{fig:no_fine_tuning_schema} shows the workflow of the system.
Firstly, we prepare the ICL examples.
The normal ICL strategy requires the CTR section, the hypothesis, and the true label.
Meanwhile, the ICL+CoT strategy requires ICL examples with reasons.
We use ChatGPT (\texttt{gpt-3.5-turbo-0613}) to generate reasoned ICL examples as it has demonstrated sufficient clinical understanding~\cite{falis2024can}.
Similar to \citet{he2023using}, We prompt ChatGPT with a phrase \emph{"Reason the answer step by step"} along with the CTR section, statement, and true label from the training dataset.
The true labels and generated explanations using the ICL strategy are then stored.
See Appendix~\ref{sec:chatgpt_hyperparams} for ChatGPT's hyperparameters used for generating explanations.

Second, we retrieve the ICL examples using either a random or BM25 retriever.
Random retriever fetches ICL examples randomly, while the BM25 retriever fetches the most similar training data to the hypothesis sentence in question.
We skip this step if we do not intend to use ICL.

Third, we choose the prompt template.
If CoT is not used, the ordinary prompt is employed.
This prompt instructs LLMs to answer using only one word, either ``Contradiction'' or ``Entailment''.
If CoT is used, the CoT prompt is used to instruct LLMs to think step by step.
Refer to \autoref{fig:no_fine_tuning_schema}.(3) and \autoref{sec:prompt_examples} for both final prompt designs.

Finally, the LLMs ingest the prompted input to generate an answer.
To obtain the prediction, we checked which label appears last in the generated answer (either \emph{``Entailment''} or \emph{``Contradiction''}).

\subsection{With Parameter Fine-tuning}
\label{sec:finetuning}
\begin{figure*}[ht]
    \centering
    \includegraphics[width=0.99\textwidth]{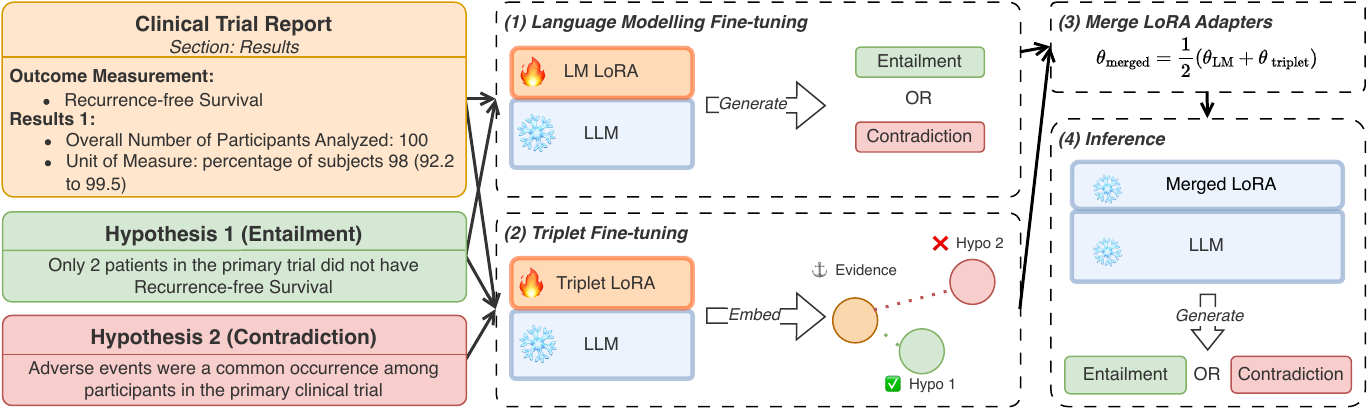}
    \caption{Our proposed fine-tuning scheme on SemEval 2024-Task 2. We suggested merging Adapters trained through Language Modelling (LM) Fine-tuning based on language modelling loss (in predicting either ``Entailment'' or ``Contradiction'') with Adapters trained through Triplet Fine-tuning based on triplet loss.}
    \label{fig:fine_tuning_schema}
\end{figure*}

We used LoRA~\cite{hu2022lora} to fine-tune the parameters $\Phi_0$ of a pretrained LLM $P_{\Phi_0}(y\mid x)$ on a training dataset $\mathcal{Z} = \{ (x_i, y_i) \}_{i=1,...,N}$.
LoRA only trains a small number of additional parameters $\theta$ 
where $|\theta| \ll |\Phi_0|$; the parameters $\theta$ introduced by LoRA are used to define a new set of parameters $\Phi$ for the LLM, such that $\Phi = \Phi_0 + \Delta \Phi(\theta)$.
The training objective for the additional parameters $\theta$ introduced by LoRA can be defined as:
\begin{equation*}  
    \argmax _{\theta} \sum_{(x, y) \in \mathcal{Z}} f\left(P_{\Phi_0 + \Delta \Phi(\theta)}\left(y \mid x\right)\right).
\end{equation*}
In our proposed method, we fine-tune two 
adapters using different training objectives, namely a Language Modelling objective (used to train the adapter parameters $\theta_{\text{LM}}$) and a supervised learning objective based on the triplet loss~\citep{DBLP:conf/bmvc/BalntasRPM16} (used to train the adapter $\theta_{\text{triplet}}$).

In the supervised learning setting, we train LLMs using a triplet loss, with CTR serving as an anchor.
Each CTR is associated with a pair of hypotheses, one contradiction and one entailment.
The triplet loss encourages LLMs to map the entailment hypothesis closer to the CTR and the contradiction hypothesis to be far from the CTR.
\begin{equation*}
L(a, p, n) = \max(0, d(a, p) - d(a, n) + \alpha),
\end{equation*}
\noindent where $a$, $p$, and $n$ denote the averaged last hidden states of the LLM for the anchor (CTR), positive sample (entailment hypothesis), and negative sample (contradiction hypothesis), respectively. $\alpha$ is a margin.
%
%

%
We hypothesise that LM fine-tuning 
can improve the accuracy of the model on knowledge-intensive domain-specific downstream tasks, while supervised fine-tuning aids the model in distinguishing syntactically similar but semantically different data points and vice versa. Merging both adapters aims to achieve the best of both fine-tuning methods:
\begin{equation*}
    \theta_{\text{merged}} = \frac{1}{2}\left( \theta_{\text{LM}} + \theta_{\text{triplet}} \right).
\end{equation*}
This process resulted in one merged LoRA adapter, which can be re-attached to the original LLM.
The base LLM, equipped with the merged LoRA, processes similarly prompted input, generating either \emph{"Entailment"} or \emph{"Contradiction"}. Refer to~\autoref{fig:fine_tuning_schema} for an illustration of the workflow.






\section{Results}

\begin{table}[t]
\centering
\resizebox{\columnwidth}{!}{
\begin{tabular}{lcccc}
\toprule
\textbf{Model} & \textbf{F1} & \textbf{Faith.} & \textbf{Con.} & \textbf{Avg.} \\
\midrule
Mistral-7B-Instruct        & 0.6525 & 0.1343 & 0.4154 & 0.4007 \\
\hspace{5mm}+ 1-shot       & 0.6639 & 0.1111 & 0.4127 & 0.3959 \\
\hspace{5mm}+ 2-shot       & 0.6685 & 0.1343 & 0.4246 & 0.4091 \\
\hspace{5mm}+ CoT          & 0.4708 & 0.5926 & 0.5077 & 0.5237 \\
\hspace{5mm}+ CoT + 1-shot & 0.5835 & 0.5706 & 0.5493 & 0.5678 \\
\hspace{5mm}+ CoT + 2-shot & 0.5944 & 0.6065 & 0.5650 & 0.5886 \\
\midrule
MistralLite-7B             & - & - & - & - \\
\hspace{5mm}+ 1-shot       & 0.5389 & 0.4109 & 0.4826 & 0.4775 \\
\hspace{5mm}+ 2-shot       & 0.4665 & 0.6597 & 0.5413 & 0.5558 \\
\hspace{5mm}+ CoT          & - & - & - & - \\
\hspace{5mm}+ CoT + 1-shot & 0.5628 & 0.4664 & 0.4973 & 0.5088 \\
\hspace{5mm}+ CoT + 2-shot & 0.5801 & 0.4977 & 0.5164 & 0.5314 \\
\midrule
LLaMA2-7B-Chat             & 0.6417 & 0.1192 & 0.4159 & 0.3923 \\
\hspace{5mm}+ 1-shot       & 0.6451 & 0.1678 & 0.4376 & 0.4168 \\
\hspace{5mm}+ 2-shot       & 0.6308 & 0.1701 & 0.4304 & 0.4104 \\
\hspace{5mm}+ CoT          & 0.6369 & 0.3009 & 0.4775 & 0.4718 \\
\hspace{5mm}+ CoT + 1-shot & 0.6101 & 0.3924 & 0.4855 & 0.4960 \\
\hspace{5mm}+ CoT + 2-shot & 0.5607 & 0.4630 & 0.4925 & 0.5054 \\
\midrule
LLaMA2-13B-Chat            & 0.6069 & 0.4502 & 0.4940 & 0.5170 \\
\hspace{5mm}+ 1-shot       & 0.6303 & 0.3345 & 0.4882 & 0.4843 \\
\hspace{5mm}+ 2-shot       & 0.6169 & 0.4016 & 0.5012 & 0.5066 \\
\hspace{5mm}+ CoT          & 0.6028 & 0.5012 & 0.5116 & 0.5385 \\
\hspace{5mm}+ CoT + 1-shot & 0.6346 & 0.5312 & 0.5360 & 0.5673 \\
\hspace{5mm}+ CoT + 2-shot & 0.5919 & 0.6123 & 0.5549 & 0.5864 \\
\midrule
\textbf{GPT-4} & \textbf{0.7751} & \textbf{0.9479} & \textbf{0.7754} & \textbf{0.8328} \\
\bottomrule
\end{tabular}%
}
\caption{Results on the test set across various LLMs with multiple prompting strategies (no fine-tuning).}
\label{table:result_no_fine_tuning}
\end{table}

The results shown in~\autoref{table:result_no_fine_tuning} can help us answer multiple research questions:

\subsection*{RQ 1: Can zero-shot LLMs perform well?}

In a zero-shot setting, MistralLite-7B showed zero performance across all metrics due to it outputting an empty string.
This suggests that, without any prompting strategies, it did not understand the given instruction.
Mistral-7B-Instruct, LLaMA2-7B-Chat, and LLaMA2-13B-Chat show some degree of performance in the F1, faithfulness, and consistency metrics.
Among the three, LLaMA2-13B-Chat achieved the highest faithfulness and consistency scores.
GPT-4 stood out with the highest scores in all metrics, suggesting its strong performance even without any prompting strategies applied.
This begs the question of whether any prompting strategies can be applied to help the relatively smaller LLMs perform better.

\subsection*{RQ 2: Can smaller LLMs perform on par with GPT-4 with prompting strategies?}

\paragraph{In-Context Learning}
We investigated 1- and 2-shot settings using BM25.
1-shot setting consistently improved the performance of the LLMs (see Appendix~\ref{sec:ablation_relevance_icl} comparing random and BM25 ICL examples).
With an ICL example, MistralLite-7B understood how to answer the prompted input.
Mistral-7B-Instruct, LLaMA2-7B-Chat, and LLaMA2-13B-Chat also showed performance improvement compared to the zero-shot setting, albeit marginal.
The 2-shot setting did not improve the LLMs consistently.
Mistral-7B-Instruct showed an improvement in all metrics with 2-shot settings, while the other LLMs see F1 score drops, albeit the faithfulness and consistency may be improved.

\paragraph{Chain-of-Thought}

We investigated CoT in a zero-shot setting.
Similarly to the zero-shot setting, MistralLite-7B showed zero performance in all metrics due to outputting an empty string.
We saw drops in F1 scores for Mistral-7B-Instruct, LLaMA2-7B-Chat, and LLaMA2-13B-Chat, and improved the faithfulness and consistency scores.
This indicates the efficacy of CoT in ensuring faithful and consistent answers from LLMs, albeit it may marginally harm the accuracy of the model.

\paragraph{In-Context Learning + Chain-of-Thought}

Since ICL improves the LLMs' F1 score, and CoT improves the faithfulness and consistency scores, we investigated the combination of both.
The results show that ICL + CoT improves LLMs across metrics.
Considering the averaged score, 2-shot ICL and CoT improve all LLMs except for MistralLite-7B.

Despite employing these strategies, the LLMs could not outperform GPT-4, particularly in terms of faithfulness and consistency.
This suggests that while combining ICL and CoT is beneficial, it is still challenging to achieve parity with GPT-4.

\subsection*{RQ 3: Can fine-tuned smaller LLMs perform on par with GPT-4?}

\begin{table}[ht]
\centering
\resizebox{\columnwidth}{!}{
\begin{tabular}{@{}lcccc@{}}
\toprule
\textbf{Model} & \textbf{F1} & \textbf{Faith.} & \textbf{Con.} & \textbf{Avg.} \\
\midrule
Mistral-7B-Instruct & \textbf{0.7689} & 0.7662 & 0.7140 & 0.7497 \\
\textbf{MistralLite-7B}      & 0.7478 & 0.8727 & \textbf{0.7220} & \textbf{0.7808} \\
LLaMA2-7B-Chat      & 0.6073 & 0.7176 & 0.6146 & 0.6465 \\
LLaMA2-13B-Chat     & 0.6766 & 0.7731 & 0.6610 & 0.7036 \\
Meditron-7B         & 0.1980 & \textbf{0.9560} & 0.6165 & 0.5902 \\
\bottomrule
\end{tabular}%
}
\caption{Results on the test set across various LLMs with parametric-efficient fine-tuning.}
\label{table:result_fine_tuning}
\end{table}

As we may have reached the limit of performance using prompting strategies, we investigated employing fine-tuning the smaller LLMs.

\paragraph{Can LoRA fine-tuning improve the performance of LLMs?}
\autoref{table:result_fine_tuning} presents the performance for each LLM fine-tuned with LoRA.
Notably, fine-tuning leads to improvements across all metrics for all LLMs.
MistralLite-7B is the best-performing LLM after fine-tuning with 0.7808 averaged scores, and it is notably better in terms of faithfulness and consistency scores compared to the other models.
The fine-tuned Meditron-7B did not show a satisfactory overall performance.
The subsequent experiment in merging LoRA adapters will focus on using MistralLite-7B as the base model.

\paragraph{Can merging LoRA adapters improve the performance of LLMs?}
\begin{table}[ht]
    \centering
    \resizebox{\columnwidth}{!}{
    \begin{tabular}{@{}lcccc@{}}
    \toprule
    \textbf{Model} & \textbf{F1} & \textbf{Faith.} & \textbf{Con.} & \textbf{Avg} \\
    \midrule
    MistralLite-7B                                         & & & & \\
    + $\theta_{\text{LM}}$                                       & 0.7478 & 0.8727 & 0.7220 & 0.7808 \\
    \textbf{+ Avg ($\theta_{\text{LM}}$, $\theta_{\text{triplet}}$)}        & \textbf{0.7824} & \textbf{0.8391} & \textbf{0.7372} & \textbf{0.7862} \\
    \bottomrule
    \end{tabular}%
    }
    \caption{Results on the test set with our proposed merging adapters fine-tuning.}
    \label{table:result_merging_peft}
\end{table}

\autoref{table:result_merging_peft} displays results obtained through fine-tuning MistralLite-7B with only LM adapter $\theta_{\text{LM}}$ and the average of $\theta_{\text{LM}}$ and $\theta_{\text{triplet}}$ adapters.
The merged $\theta_{\text{LM}}$ and $\theta_{\text{triplet}}$ adapters improve the overall performance of the LLM (joint-fourth in the competition).
It achieves a better F1 score of 0.7824 (+0.0346), indicating that merging LoRA adapters may improve the predictive performance of LLMs.
We noticed a lower faithfulness score (-0.0336) and a higher consistency score (+0.0152).
This indicates the model struggles to understand semantic changes introduced by deliberate alterations but can understand semantically similar data better.



\subsection{Contamination Analysis on GPT-4}

Inspired by \citet{carlini2022quantifying},
we assessed
whether instances of the NLI4CT dataset were included in GPT-4's pre-training data.
We prompted GPT-4 with:
1) System instruction: \textit{"You are a helpful assistant on the SemEval task. Complete the given statement."},
2) Truncation of half of the statement to prompt GPT-4 to infer the remaining. (refer to Appendices \ref{sec:contamination_hyperparams} and \ref{sec:example_gpt4_contamination} for details)

We define two metrics: \textit{extractable match}, checking if the predicted half of the statement by GPT-4 is included in the original half, and \textit{partial match}, assessing how sequentially each token of the predicted half of the statement is included in the original half.
In the test set, GPT-4 recorded an extractable match score of 0.033 and a partial match score of 0.322.
The low extractable match score may indicate that GPT-4 has not seen the test data during its pretraining, whereas the higher partial match score may indicate GPT-4's ability to identify keywords from CTRs.

\section{Conclusion}

This study assesses the performance of various LLMs, 
employing diverse strategies such as CoT, ICL, and PEFT.
We propose a PEFT method, merging independent adapters fine-tuned separately using triplet and LM losses.
Our proposed PEFT method improves the F1 and consistency scores but reduces faithfulness --- our best fine-tuned model, MistralLite-7B + LM LoRA + Triplet LoRA, achieved an average score of 0.7862.
However, it does not outperform GPT-4 in terms of faithfulness and consistency: GPT-4 ranks joint-first in the competition with an average score of 0.8328.
A contamination analysis on GPT-4 revealed no NLI4CT test data leakage, indicated by a low extractable match score (0.033), and showcased its ability to identify keywords from CTRs with a relatively high partial match score (0.322).

\section*{Limitations}

Due to the scope of the study and the limited resources, we opted to only experiment with GPT-4 in a zero-shot setup.
However, our proposed strategies that improved the performance of smaller LLMs could also be used to enhance GPT-4.
Albeit the promising performance of the LLMs, particularly GPT-4, the predictions may still be inaccurate and should not be used in a clinical setting without human supervision.

We conducted a contamination analysis inspired by~\citet{carlini2022quantifying} and concluded that there may be no test data leakage during the pretraining of GPT-4.
However, we acknowledge that contamination analysis alone may not be sufficient in proving test data leakage.


\section*{Acknowledgements}
APG was supported by the United Kingdom Research and Innovation (grant EP/S02431X/1), UKRI Centre for Doctoral Training in Biomedical AI at the University of Edinburgh, School of Informatics. 
PM was partially funded by 
ELIAI (The Edinburgh Laboratory for Integrated Artificial Intelligence), EPSRC (grant no. EP/W002876/1), an industry grant from Cisco, and a donation from Accenture LLP; and is grateful to NVIDIA for the GPU donations.
BA was partially funded by Legal and General PLC as part of the Advanced Care Research Centre and by the Artificial Intelligence and Multimorbidity: Clustering in Individuals, Space and Clinical Context (AIM-CISC) grant NIHR202639.
For the purpose of open access, The authors have applied a Creative Commons attribution (CC BY) licence to any author-accepted manuscript version arising.
Experiments from this work are conducted mainly on the Edinburgh International Data Facility\footnote{https://edinburgh-international-data-facility.ed.ac.uk/} and supported by the Data-Driven Innovation Programme at the University of Edinburgh.


\bibliography{custom}

\begin{thebibliography}{22}
\expandafter\ifx\csname natexlab\endcsname\relax\def\natexlab#1{#1}\fi

\bibitem[{Balntas et~al.(2016)Balntas, Riba, Ponsa, and Mikolajczyk}]{DBLP:conf/bmvc/BalntasRPM16}
Vassileios Balntas, Edgar Riba, Daniel Ponsa, and Krystian Mikolajczyk. 2016.
\newblock Learning local feature descriptors with triplets and shallow convolutional neural networks.
\newblock In \emph{{BMVC}}. {BMVA} Press.

\bibitem[{Carlini et~al.(2022)Carlini, Ippolito, Jagielski, Lee, Tramer, and Zhang}]{carlini2022quantifying}
Nicholas Carlini, Daphne Ippolito, Matthew Jagielski, Katherine Lee, Florian Tramer, and Chiyuan Zhang. 2022.
\newblock Quantifying memorization across neural language models.
\newblock In \emph{The Eleventh International Conference on Learning Representations}.

\bibitem[{Chen et~al.(2021)Chen, Tworek, Jun, Yuan, Pinto, Kaplan, Edwards, Burda, Joseph, Brockman et~al.}]{chen2021evaluating}
Mark Chen, Jerry Tworek, Heewoo Jun, Qiming Yuan, Henrique Ponde de~Oliveira Pinto, Jared Kaplan, Harri Edwards, Yuri Burda, Nicholas Joseph, Greg Brockman, et~al. 2021.
\newblock Evaluating large language models trained on code.
\newblock \emph{arXiv preprint arXiv:2107.03374}.

\bibitem[{Chronopoulou et~al.(2023)Chronopoulou, Peters, Fraser, and Dodge}]{chronopoulou2023adaptersoup}
Alexandra Chronopoulou, Matthew~E Peters, Alexander Fraser, and Jesse Dodge. 2023.
\newblock Adaptersoup: Weight averaging to improve generalization of pretrained language models.
\newblock \emph{arXiv preprint arXiv:2302.07027}.

\bibitem[{Falis et~al.(2024)Falis, Gema, Dong, Daines, Basetti, Holder, Penfold, Birch, and Alex}]{falis2024can}
Mat{\'u}{\v{s}} Falis, Aryo~Pradipta Gema, Hang Dong, Luke Daines, Siddharth Basetti, Michael Holder, Rose~S Penfold, Alexandra Birch, and Beatrice Alex. 2024.
\newblock Can gpt-3.5 generate and code discharge summaries?
\newblock \emph{arXiv preprint arXiv:2401.13512}.

\bibitem[{Gema et~al.(2023)Gema, Daines, Minervini, and Alex}]{gema2023parameter}
Aryo~Pradipta Gema, Luke Daines, Pasquale Minervini, and Beatrice Alex. 2023.
\newblock Parameter-efficient fine-tuning of llama for the clinical domain.
\newblock \emph{arXiv preprint arXiv:2307.03042}.

\bibitem[{He et~al.(2023)He, Wu, Camburu, Minervini, and Stenetorp}]{he2023using}
Xuanli He, Yuxiang Wu, Oana-Maria Camburu, Pasquale Minervini, and Pontus Stenetorp. 2023.
\newblock Using natural language explanations to improve robustness of in-context learning for natural language inference.
\newblock \emph{arXiv preprint arXiv:2311.07556}.

\bibitem[{Hu et~al.(2022)Hu, Shen, Wallis, Allen-Zhu, Li, Wang, Wang, and Chen}]{hu2022lora}
Edward~J Hu, Yelong Shen, Phillip Wallis, Zeyuan Allen-Zhu, Yuanzhi Li, Shean Wang, Lu~Wang, and Weizhu Chen. 2022.
\newblock \href {https://openreview.net/forum?id=nZeVKeeFYf9} {Lo{RA}: Low-rank adaptation of large language models}.
\newblock In \emph{International Conference on Learning Representations}.

\bibitem[{Jiang et~al.(2023)Jiang, Sablayrolles, Mensch, Bamford, Chaplot, Casas, Bressand, Lengyel, Lample, Saulnier et~al.}]{jiang2023mistral}
Albert~Q Jiang, Alexandre Sablayrolles, Arthur Mensch, Chris Bamford, Devendra~Singh Chaplot, Diego de~las Casas, Florian Bressand, Gianna Lengyel, Guillaume Lample, Lucile Saulnier, et~al. 2023.
\newblock Mistral 7b.
\newblock \emph{arXiv preprint arXiv:2310.06825}.

\bibitem[{Jullien et~al.(2024)Jullien, Valentino, and Freitas}]{jullien-etal-2024-semeval}
Ma{\"e}l Jullien, Marco Valentino, and Andr{\'e} Freitas. 2024.
\newblock {S}em{E}val-2024 task 2: Safe biomedical natural language inference for clinical trials.
\newblock In \emph{Proceedings of the 18th International Workshop on Semantic Evaluation (SemEval-2024)}. Association for Computational Linguistics.

\bibitem[{Jullien et~al.(2023{\natexlab{a}})Jullien, Valentino, Frost, O{'}Regan, Landers, and Freitas}]{jullien-etal-2023-nli4ct}
Mael Jullien, Marco Valentino, Hannah Frost, Paul O{'}Regan, D{\'o}nal Landers, and Andre Freitas. 2023{\natexlab{a}}.
\newblock \href {https://doi.org/10.18653/v1/2023.emnlp-main.1041} {{NLI}4{CT}: Multi-evidence natural language inference for clinical trial reports}.
\newblock In \emph{Proceedings of the 2023 Conference on Empirical Methods in Natural Language Processing}, pages 16745--16764, Singapore. Association for Computational Linguistics.

\bibitem[{Jullien et~al.(2023{\natexlab{b}})Jullien, Valentino, Frost, O{'}regan, Landers, and Freitas}]{jullien-etal-2023-semeval}
Ma{\"e}l Jullien, Marco Valentino, Hannah Frost, Paul O{'}regan, Donal Landers, and Andr{\'e} Freitas. 2023{\natexlab{b}}.
\newblock \href {https://doi.org/10.18653/v1/2023.semeval-1.307} {{S}em{E}val-2023 task 7: Multi-evidence natural language inference for clinical trial data}.
\newblock In \emph{Proceedings of the 17th International Workshop on Semantic Evaluation (SemEval-2023)}, pages 2216--2226, Toronto, Canada. Association for Computational Linguistics.

\bibitem[{Kojima et~al.(2022)Kojima, Gu, Reid, Matsuo, and Iwasawa}]{kojima2022large}
Takeshi Kojima, Shixiang~Shane Gu, Machel Reid, Yutaka Matsuo, and Yusuke Iwasawa. 2022.
\newblock Large language models are zero-shot reasoners.
\newblock \emph{Advances in neural information processing systems}, 35:22199--22213.

\bibitem[{Lehman et~al.(2023)Lehman, Hernandez, Mahajan, Wulff, Smith, Ziegler, Nadler, Szolovits, Johnson, and Alsentzer}]{lehman2023we}
Eric Lehman, Evan Hernandez, Diwakar Mahajan, Jonas Wulff, Micah~J Smith, Zachary Ziegler, Daniel Nadler, Peter Szolovits, Alistair Johnson, and Emily Alsentzer. 2023.
\newblock Do we still need clinical language models?
\newblock \emph{arXiv preprint arXiv:2302.08091}.

\bibitem[{Li{\'e}vin et~al.(2022)Li{\'e}vin, Hother, and Winther}]{lievin2022can}
Valentin Li{\'e}vin, Christoffer~Egeberg Hother, and Ole Winther. 2022.
\newblock Can large language models reason about medical questions?
\newblock \emph{arXiv preprint arXiv:2207.08143}.

\bibitem[{Mangrulkar et~al.(2022)Mangrulkar, Gugger, Debut, Belkada, and Paul}]{peft}
Sourab Mangrulkar, Sylvain Gugger, Lysandre Debut, Younes Belkada, and Sayak Paul. 2022.
\newblock Peft: State-of-the-art parameter-efficient fine-tuning methods.
\newblock \url{https://github.com/huggingface/peft}.

\bibitem[{OpenAI(2023)}]{openai2023gpt4}
OpenAI. 2023.
\newblock \href {https://doi.org/10.48550/ARXIV.2303.08774} {{GPT-4} technical report}.
\newblock \emph{CoRR}, abs/2303.08774.

\bibitem[{Ouyang et~al.(2022)Ouyang, Wu, Jiang, Almeida, Wainwright, Mishkin, Zhang, Agarwal, Slama, Ray et~al.}]{ouyang2022training}
Long Ouyang, Jeffrey Wu, Xu~Jiang, Diogo Almeida, Carroll Wainwright, Pamela Mishkin, Chong Zhang, Sandhini Agarwal, Katarina Slama, Alex Ray, et~al. 2022.
\newblock Training language models to follow instructions with human feedback.
\newblock \emph{Advances in Neural Information Processing Systems}, 35:27730--27744.

\bibitem[{Touvron et~al.(2023{\natexlab{a}})Touvron, Lavril, Izacard, Martinet, Lachaux, Lacroix, Rozi{\`e}re, Goyal, Hambro, Azhar et~al.}]{touvron2023allama}
Hugo Touvron, Thibaut Lavril, Gautier Izacard, Xavier Martinet, Marie-Anne Lachaux, Timoth{\'e}e Lacroix, Baptiste Rozi{\`e}re, Naman Goyal, Eric Hambro, Faisal Azhar, et~al. 2023{\natexlab{a}}.
\newblock Llama: Open and efficient foundation language models.
\newblock \emph{arXiv preprint arXiv:2302.13971}.

\bibitem[{Touvron et~al.(2023{\natexlab{b}})Touvron, Martin, Stone, Albert, Almahairi, Babaei, Bashlykov, Batra, Bhargava, Bhosale et~al.}]{touvron2023bllama}
Hugo Touvron, Louis Martin, Kevin Stone, Peter Albert, Amjad Almahairi, Yasmine Babaei, Nikolay Bashlykov, Soumya Batra, Prajjwal Bhargava, Shruti Bhosale, et~al. 2023{\natexlab{b}}.
\newblock Llama 2: Open foundation and fine-tuned chat models.
\newblock \emph{arXiv preprint arXiv:2307.09288}.

\bibitem[{Wolf et~al.(2020)Wolf, Debut, Sanh, Chaumond, Delangue, Moi, Cistac, Rault, Louf, Funtowicz, Davison, Shleifer, von Platen, Ma, Jernite, Plu, Xu, Scao, Gugger, Drame, Lhoest, and Rush}]{wolf-etal-2020-transformers}
Thomas Wolf, Lysandre Debut, Victor Sanh, Julien Chaumond, Clement Delangue, Anthony Moi, Pierric Cistac, Tim Rault, Rémi Louf, Morgan Funtowicz, Joe Davison, Sam Shleifer, Patrick von Platen, Clara Ma, Yacine Jernite, Julien Plu, Canwen Xu, Teven~Le Scao, Sylvain Gugger, Mariama Drame, Quentin Lhoest, and Alexander~M. Rush. 2020.
\newblock \href {https://www.aclweb.org/anthology/2020.emnlp-demos.6} {Transformers: State-of-the-art natural language processing}.
\newblock In \emph{Proceedings of the 2020 Conference on Empirical Methods in Natural Language Processing: System Demonstrations}, pages 38--45, Online. Association for Computational Linguistics.

\bibitem[{{Yin Song and Chen Wu and Eden Duthie}(2023)}]{MistralLite_2023}
{Yin Song and Chen Wu and Eden Duthie}. 2023.
\newblock \href {https://huggingface.co/amazon/MistralLite} {{amazon/MistralLite}}.

\end{thebibliography}

\clearpage

\appendix

\section{Experimental setup}
\label{sec:app_setup}
We use HuggingFace's Transformers~\cite{wolf-etal-2020-transformers} and PEFT~\cite{peft} libraries for the experiments.
All inferences and fine-tuning experiments were run on two NVIDIA A100-40GB GPUs.

For models without parameter fine-tuning (prompting strategies, \autoref{sec:no_finetuning}), in-context examples were retrieved from the Training set (for both random and BM25 retrievers). Additionally, the Dev set was used to evaluate and select the optimal prompt design. Models with parameter fine-tuning (\autoref{sec:finetuning}) were trained using the Training set, and the Dev set was utilised to determine the best checkpoint.

\section{Hyperparameters}
\label{sec:app_hyperparams}

\subsection{ChatGPT Hyperparameters for the generation of Natural Language Explanation}
\label{sec:chatgpt_hyperparams}

We prompted GPT-3.5 (model name: \texttt{gpt-3.5-turbo-0613}) with hyperparameters as shown in~\autoref{tab:api_hyperparameters}.
The generation process took approximately 2 hours and cost \$2.

\begin{table}
    \centering
    \small
    \begin{tabular}{lp{0.6\linewidth}}
    \toprule
    {\bf Parameter} & {\bf Value} \\
    \midrule
    Model Name        & gpt-3.5-turbo-0613 \\
    API Version       & 2023-03-15-preview \\
    Temperature       & 0 \\
    Top P             & 0 \\
    Frequency Penalty & 0 \\
    Presence Penalty  & 0 \\
    Max new token     & 256 \\
    System Prompt     & You are a helpful clinician's assistant designed to identify if a clinical statement is a contradiction or an entailment to the presented evidence. \\
    Prompt            & Evidence: [Evidence] \newline Statement: [Statement] \newline Question: Answer in 1 word. Is the statement a contradiction or an entailment? \newline Answer: [Label] \newline Reason the answer step by step \\
    \bottomrule
    \end{tabular}
    \caption{Azure API call hyperparameters.}
    \label{tab:api_hyperparameters}
\end{table}

\subsection{GPT-4 generation hyperparameters}
\label{sec:gpt4_gen_hyperparams}

We prompted GPT-4 (model name: \texttt{gpt-4}) with the ordinary prompt as shown in~\autoref{fig:no_fine_tuning_schema}.
We set \texttt{temperature=0} to ensure that the model's generation is deterministic.
The maximum generation length is 8.
The generation process took approximately 2 hours and cost \$77.

\subsection{Non GPT-4 generation hyperparameters}
\label{sec:non_gpt4_gen_hyperparams}

All models (apart from GPT-4) were loaded in BFloat16 to ensure that they fit into our resources.
We used \texttt{do\_sample=False} to ensure that the model's generation is deterministic.
The maximum generation length is 8 new tokens for non-CoT experiments and 100 for CoT experiments. 

\subsection{Language Modelling training hyperparameters}
\label{sec:lm_hyperparams}

\begin{table}[t]
    \centering
    \small
    \begin{tabular}{lc}
        \toprule
        Hyperparameter & Value \\
        \midrule
        Epoch & 10 \\
        Gradient accumulation step & 32 \\
        Optimiser & AdamW \\
        Learning rate & 0.001 \\
        Weight decay & 0.01 \\
        Max sequence length & 2048 \\
        \bottomrule
    \end{tabular}
    \caption{Language Modelling training hyperparameters.}
    \label{tab:lm_train_hyperparams}
\end{table}

LM training used the hyperparameters detailed in~\autoref{tab:lm_train_hyperparams}.
The LLM's maximum sequence length is adjusted to fit on two NVIDIA A100-40GB GPUs.

\subsection{Triplet training hyperparameters}
\label{sec:triplet_hyperparams}

\begin{table}[t]
    \centering
    \small
    \begin{tabular}{lc}
        \toprule
        Hyperparameter & Value \\
        \midrule
        Epoch & 10 \\
        Gradient accumulation step & 32 \\
        Optimiser & AdamW \\
        Learning rate & 0.00001 \\
        Weight decay & 0.01 \\
        Max sequence length & 1024 \\
        Triplet loss margin & 1.0 \\
        Triplet loss $p$ & 2 \\
        Triplet loss $\epsilon$ & 1e-7 \\
        \bottomrule
    \end{tabular}
    \caption{Triplet training hyperparameters.}
    \label{tab:triplet_train_hyperparams}
\end{table}

Triplet training used the hyperparameters detailed in~\autoref{tab:triplet_train_hyperparams}.
The LLM's maximum sequence length is adjusted to fit on two NVIDIA A100-40GB GPUs.
Triplet training demands more memory because we need to generate three hidden representations during training (\ie anchor, positive, negative), necessitating a reduction in sequence length.

\subsection{PEFT Hyperparameters}
\label{sec:peft_hyperparams}

\begin{table}[t]
    \centering
    \small
    \begin{tabular}{lc}
        \toprule
        Hyperparameter & Value \\
        \midrule
        r & 16 \\
        alpha & 32 \\
        dropout & 0.0 \\
        target\_modules & [``k\_proj'', ``q\_proj'', ``v\_proj''] \\
        \bottomrule
    \end{tabular}
    \caption{LoRA Hyperparameters.}
    \label{tab:lora_hyperparams}
\end{table}

All LLMs and training methods (\ie LM and triplet training) used the same LoRA hyperparameters as shown in~\autoref{tab:lora_hyperparams}.

\subsection{Contamination Analysis on GPT-4}
\label{sec:contamination_hyperparams}

For the Contamination Analysis, we utilised the same settings as those described in Appendix \ref{sec:gpt4_gen_hyperparams}, specifically setting the maximum number of generated tokens to 8. This was done to prevent the incorrect biases due to excessively lengthy predictions by GPT-4, as our evaluation method focuses on determining whether the prediction is included within the ground truth.

\section{Ablation study on Random vs Relevance-based In-Context Examples}
\label{sec:ablation_relevance_icl}

\begin{table}[t]
\centering
\resizebox{\columnwidth}{!}{
\begin{tabular}{llcccc}
\toprule
\textbf{Model} & ICL & \textbf{F1} & \textbf{Faith.} & \textbf{Con.} & \textbf{Avg.} \\
\midrule
Mistral-7b-Instruct & Random: 1-shot & 0.6694 & 0.0856 & 0.4086 & 0.3879 \\
Mistral-7b-Instruct & BM25: 1-shot & 0.6639 & 0.1111 & 0.4127 & 0.3959 \\
Mistral-7b-Instruct & Random: 2-shot & 0.6639 & 0.1458 & 0.4294 & 0.4130 \\
Mistral-7b-Instruct & BM25: 2-shot & 0.6685 & 0.1343 & 0.4246 & 0.4091 \\
\midrule
MistralLite-7B & Random: 1-shot & 0.6622 & 0.0150 & 0.3854 & 0.3542 \\
MistralLite-7B & BM25: 1-shot & 0.5389 & 0.4109 & 0.4826 & 0.4775 \\
MistralLite-7B & Random: 2-shot & 0.5097 & 0.5023 & 0.5164 & 0.5095 \\
MistralLite-7B & BM25: 2-shot & 0.4665 & 0.6597 & 0.5413 & 0.5558 \\
\midrule
LLaMA2-7B-Chat & Random: 1-shot & 0.6613 & 0.0116 & 0.3864 & 0.3531 \\
LLaMA2-7B-Chat & BM25: 1-shot & 0.6451 & 0.1678 & 0.4376 & 0.4168 \\
LLaMA2-7B-Chat & Random: 2-shot & 0.6387 & 0.1250 & 0.4180 & 0.3939 \\
LLaMA2-7B-Chat & BM25: 2-shot & 0.6308 & 0.1701 & 0.4304 & 0.4104 \\
\midrule
LLaMA2-13B-Chat & Random: 1-shot & 0.6585 & 0.3113 & 0.4724 & 0.4807 \\
LLaMA2-13B-Chat & BM25: 1-shot & 0.6303 & 0.3345 & 0.4882 & 0.4843 \\
LLaMA2-13B-Chat & Random: 2-shot & 0.6230 & 0.4074 & 0.4935 & 0.5080 \\
LLaMA2-13B-Chat & BM25: 2-shot & 0.6169 & 0.4016 & 0.5012 & 0.5066 \\
\bottomrule
\end{tabular}%
}
\caption{Comparison of In-Context Learning Models Using Random and BM25 Retrievers on the Test set}
\label{table:ablation_relevance_icl}
\end{table}

We also compared the performance of the model by using random and relevant ICL examples.
As shown in~\autoref{table:ablation_relevance_icl}, we found that relevant ICL examples helped the LLMs achieve better faithfulness and consistency scores, while the F1 scores may be impacted.
For that reason, we opted to use relevance-based ICL examples for the ICL-based runs.






\section{Prompt Examples}
\label{sec:prompt_examples}
Here, we provide examples of the prompts used in our experiments. The black text within the box represents the prompt input text, the \textcolor{red}{red} text represents the prediction of the models, and the \textcolor{blue}{blue} text represents the ground truth.

\subsection{Base Prompt for Zero-shot and In-Context Learning}
\label{sec:example_zero}

\begin{tcolorbox}[colback=gray!5!white,colframe=blue!75!black,colbacktitle=gray!80!black, title=Base Prompt,fonttitle=\bfseries]
You are a helpful assistant.\\

<<In-Context Learning Examples>>\\

Evidence: \\
Primary trial:\\
Adverse Events 1:\\
Total: 0/15 (0.00\%)\\
Adverse Events 2:\\
Total: \\\\
Secondary trial:\\
Adverse Events 1:\\
Total: 0/442 (0.00\%)\\
Adverse Events 2:\\\\
Statement: the primary trial and the secondary trial do not have any recorded adverse events for their participants. crypt is a pitlike depression or tubular recess.\\
Question: Answer in 1 word. Is the statement a contradiction or an entailment?\\
Answer: \textcolor{red}{Entailment}
\\\\
\textcolor{blue}{Ground Truth:  Entailment}
\end{tcolorbox}

\newpage

\subsection{Chain-of-Thought for Zero-shot and In-Context Learning}
\label{sec:example_CoT}

\begin{tcolorbox}[colback=gray!5!white,colframe=blue!75!black,colbacktitle=gray!80!black, title=Chain-of-Thought,fonttitle=\bfseries]
You are a helpful assistant.\\

<<In-Context Learning Examples>>\\

Evidence: \\
Primary trial:\\
Adverse Events 1:\\
Total: 0/15 (0.00\%)\\
Adverse Events 2:\\
Total: \\\\
Secondary trial:\\
Adverse Events 1:\\
Total: 0/442 (0.00\%)\\
Adverse Events 2:\\\\
Statement: the primary trial and the secondary trial do not have any recorded adverse events for their participants. crypt is a pitlike depression or tubular recess.\\
Question: Is the statement a contradiction or an entailment?\\
Let's think step by step\\
Answer: \textcolor{red}{ Great, let's analyze the statement and the evidence provided to determine if it's ... because the evidence shows that there are no adverse events recorded for the participants in either trial.\\
Therefore, the answer is Entailment."}
\\\\
\textcolor{blue}{Ground Truth:  Entailment}
\end{tcolorbox}

\newpage

\subsection{Contamination Analysis on GPT-4}
\label{sec:example_gpt4_contamination}



\begin{tcolorbox}[colback=gray!5!white,colframe=blue!75!black,colbacktitle=gray!80!black, title=Extractable Match,fonttitle=\bfseries]
You are a helpful assistant on the semeval task. Complete the given statement.\\

Evidence: \\
Primary trial:\\
Outcome Measurement: \\
Number of Participants With Reduction in CTCs Following High-dose Chemotherapy With Purged Autologous Stem Cell Products\\
Number of circulating tumor cells (CTCs) measured at one month post autologous hematopoietic stem cell transplantation (AHST), considered both as longitudinal values and compared to the baseline number of CTCs.\\
Time frame: Baseline to 1 month post AHST\\
Results 1: \\
Arm/Group Title: High-dose Chemotherapy\\
Arm/Group Description: Carboplatin + Cyclophosphamide + Thiotepa\\
Carboplatin : Target AUC of 20, then divided into 4 doses given by vein (IV) days -6, -5, -4, -3 prior to stem cell infusion.\\
Thiotepa : $120 mg/m^2$ by vein days -6, -5, -4, -3 prior to stem cell infusion.\\
Stem Cell Transplant : Stem Cell Transplant on Day 0.\\
Cyclophosphamide : $1.5 gm/m^2$ by vein days -6, -5, -4, -3 prior to stem cell infusion.\\
Overall Number of Participants Analyzed: 21\\
Measure Type: Number\\
Unit of Measure: participants  9\\
Statement: less than half of the primary trial participants had a Reduction in circulating tumor cells \textcolor{red}{Following High-dose Chemotherapy With Pur}
\\\\
\textcolor{blue}{Ground Truth:} \textcolor{red}{Following High-dose Chemotherapy With Pur}\textcolor{blue}{ged Autologous Stem Cell Products}
\end{tcolorbox}

\newpage

\begin{tcolorbox}[colback=gray!5!white,colframe=blue!75!black,colbacktitle=gray!80!black, title=Partial Match,fonttitle=\bfseries]
You are a helpful assistant on the semeval task. Complete the given statement.\\

Evidence: \\
Primary trial:\\
Adverse Events 1:\\
Total: 3/12 (25.00\%)\\
Hemoglobin 1/12 (8.33\%)\\  
Alkaline phosphatase 1/12 (8.33\%)\\
Dehydration 1/12 (8.33\%)\\
Syncope 2/12 (16.67\%)\\
Dyspnea 1/12 (8.33\%)\\
Hypotension 1/12 (8.33\%)\\\\
Secondary trial:\\
Adverse Events 1:\\
Total: 0/115 (0.00\%)\\
Deep vein thrombosis * [1]0/115 (0.00\%)\\
Adverse Events 2:\\
Total: 1/119 (0.84\%)\\
Deep vein thrombosis * [1]1/119 (0.84\%)\\
Statement: on both the primary and secondary clinical trials, syncope \textcolor{red}{was reported as an adverse event in the}
\\\\
\textcolor{blue}{Ground Truth: emerged \textcolor{red}{as} the most common} \textcolor{red}{adverse} \textcolor{blue}{occurrence} \textcolor{red}{in the} \textcolor{blue}{patient groups}
\end{tcolorbox}

\end{document}